\UseRawInputEncoding
\documentclass[conference]{IEEEtran}
\IEEEoverridecommandlockouts
\usepackage{cite}
\usepackage{amsmath,amssymb,amsfonts}
\usepackage{algorithmic}
\usepackage{graphicx}
\usepackage{textcomp}
\usepackage{xcolor}
\usepackage{verbatim}
\usepackage{float} 
\usepackage{textcomp}
\usepackage{amssymb}
\usepackage[numbers,sort&compress]{natbib}

\def\BibTeX{{\rm B\kern-.05em{\sc i\kern-.025em b}\kern-.08em
    T\kern-.1667em\lower.7ex\hbox{E}\kern-.125emX}}

\usepackage{tikz,xcolor,hyperref}

\definecolor{lime}{HTML}{A6CE39}
\DeclareRobustCommand{\orcidicon}{%
	\begin{tikzpicture}
	\draw[lime, fill=lime] (0,0) 
	circle [radius=0.16] 
	node[white] {{\fontfamily{qag}\selectfont \tiny ID}};
	\draw[white, fill=white] (-0.0625,0.095) 
	circle [radius=0.007];
	\end{tikzpicture}
	\hspace{-2mm}
}

\foreach \x in {A, ..., Z}{%
	\expandafter\xdef\csname orcid\x\endcsname{\noexpand\href{https://orcid.org/\csname orcidauthor\x\endcsname}{\noexpand\orcidicon}}
}

\begin{document}

\title{Effect of Human Involvement on Work Performance and Fluency in Human-Robot Collaboration for Recycling\\
}

\author{\IEEEauthorblockN{Sruthi Ramadurai}
\IEEEauthorblockA{\textit{Mechanical and Industrial Engineering} \\
\textit{University of Illinois at Chicago}\\
Chicago, USA \\
sramad5@uic.edu \orcidA{}}

\and
\IEEEauthorblockN{Heejin Jeong}
\IEEEauthorblockA{\textit{Mechanical and Industrial Engineering} \\
\textit{University of Illinois at Chicago}\\
Chicago, USA \\
heejinj@uic.edu \orcidB{}}
}



\maketitle

\begin{abstract}
Human-robot collaboration has significant potential in recycling due to the wide variation in the composition of recyclable products. Six participants performed a recyclable item sorting task collaborating with a robot arm equipped with a vision system. The effect of three different levels of human involvement or assistance to the robot (Level 1- occlusion removal; Level 2- optimal spacing; Level 3- optimal grip) on performance metrics such as robot accuracy, task time and subjective fluency were assessed. Results showed that human involvement had a remarkable impact on the robot’s accuracy, which increased with human involvement level. Mean accuracy values were 33.3\% for Level 1, 69\% for Level 2 and 100\% for Level 3. The results imply that for sorting processes involving diverse materials that vary in size, shape, and composition, human assistance could improve the robot’s accuracy to a significant extent while also being cost-effective.

\end{abstract}


\section{Introduction}
The fourth industrial revolution or Industry 4.0 refers to a digital revolution in industrial production as it goes through a process of change towards flexible and intelligent manufacturing, where people, machines, and products are directly connected with each other and their environment \cite{Schwab2017,Matt2020,Gomez2017}. Human-robot collaboration (HRC) is one of the key elements of Industry 4.0 \cite{Chacon2021,Choi2022}. Collaborative robots (cobots) are designed with safety features to work with humans in a shared space and achieve a common goal. While robots have the strengths of high precision, speed, repeatability, and handling of toxic materials, humans have improved cognitive skills, recognition, creative decision-making, and learned experience \cite{Villani2018}. HRC involves making the best use of both human and robot abilities to create better outcomes such as overall cost reduction, improved process efficiency, and enhanced human workers' well-being. In the present age of sustainable development, where the thrust is given to achieving a circular economy, a growing body of literature indicates that HRC has significant potential in recycling and remanufacturing \cite{Renteria2019,Mohammed2021}.

The need for human-robot collaboration in recycling stems primarily from the fact that there is wide variation in the composition of products to be recycled, such as electronic devices, household appliances, plastic consumer goods and metallic products. They consist of both valuable and hazardous materials that need to be intelligently segregated. The sorting (or classification) of materials and products is an essential task in the recycling process, because the extracted material's quality and value depend on it. Although an experienced human worker could perform the sorting task manually, there are several health risks due to exposure to hazardous materials, especially in electronics, and other contaminants present in the waste stream \cite{Poole2017} (See an example in Figure 1-(a)). On the other hand, there are challenges in automating the sorting process due to the lack of uniformity of products \cite{Renteria2019}. (See Figure 1-(b)). In such scenarios, a collaboration between humans and robots is the optimal solution (See Figure 1-(c)). Sarc et al. \cite{Sarc2019} have done a detailed review of intelligent robotics in circular economy-oriented waste management, including a discussion of commercially available fully automated waste sorting robots. Examples include Apple’s Liam and Daisy (metal recycling robots), ZenRobotics’ Fast Picker (which sorts municipal solid waste), BHS’s MAX AI AQc (which can recognize recyclables and pick them up from the conveyor belt), and Sorting Systems Bollegraaf (which can sort different types of plastic, paper, cardboard and cardboard packaging). The challenges of using robotics to sort waste include the heterogeneity of products, surface contamination of the waste stream, varying shapes or masses that the robotic arm must grab, random location of objects in the waste stream on the conveyor belt, and failure of the robot arm to grab the object owing to position changes due to vibration of the conveyor belt \cite{Sarc2019}. The practical limitations of fully automated waste sorters include the high cost of the technology and lower efficiency and dexterity compared to a human. For the sorting of different types of plastics, for instance, optical recognition still faces challenges with new bio-sourced materials. In such cases, human intervention and learned experience are not yet irreplaceable \cite{Guyot-Phung2019}. Furthermore, HRC solutions are more cost-effective compared to fully automated solutions.

\begin{figure*}
  \includegraphics[width=1.0\textwidth]{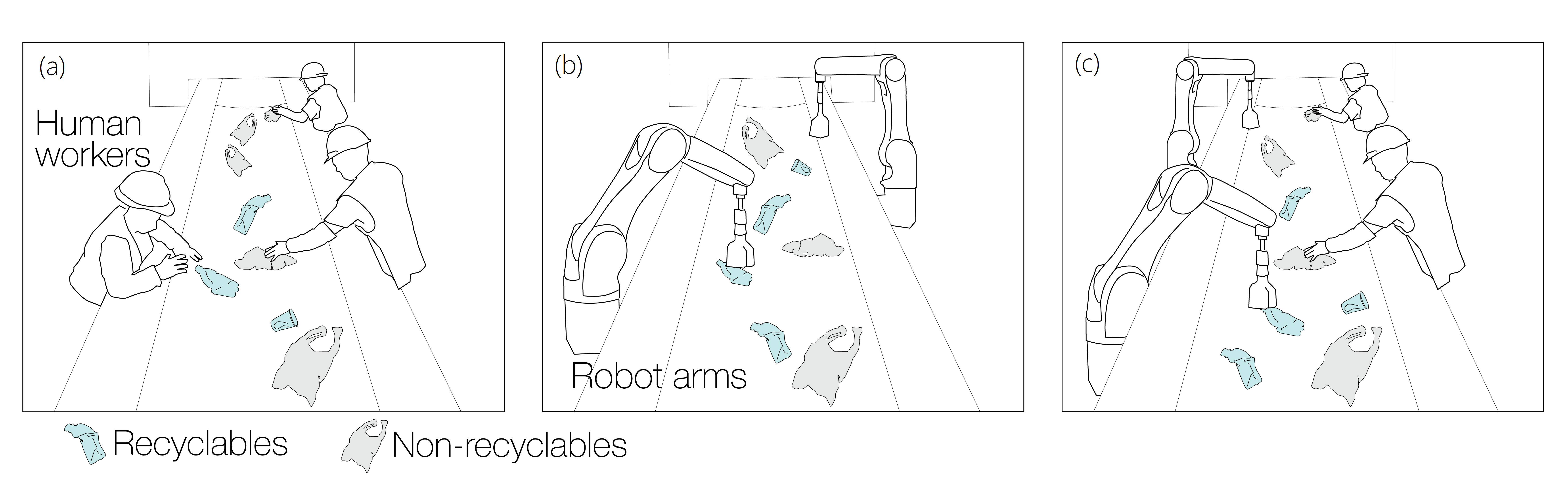}
  \vspace{-28pt}
  \caption{Illustration of Human-Robot Collaboration in a recycling task.}
\end{figure*}

Axenopοulos et al. \cite{Axeno2019} have discussed a novel human-robot collaboration framework for electrical and electronic waste recycling. They propose a hybrid human and robot collaboration approach for device classification, disassembly, and component sorting in electronic waste recycling. However, no studies investigated human-robot collaboration for waste sorting or material classification scenarios using rigorous experiments in a controlled environment to the best of our knowledge. We wanted to investigate the efficiency of human-robot collaboration in sorting lines at recycling plants (Figure 1-(c)), which currently have either only human workers (Figure 1-(a)) or only fully automated sorting robots (Figure 1-(b)). Our objective was to evaluate HRC for a plastic recycling setting, which was simplified to make its performance feasible in the laboratory. Our specific aim was to investigate the effect of human involvement on task performance and fluency for a recyclable item sorting task. As a secondary factor, we also investigated the effect of different proportions of recyclables on the same dependent variables. 

\section{Methods}

\subsection{Participants}
Six participants (5 males and 1 female, age = 23.3\textpm  0.5 y.o.) were recruited for the experiment. The subjects were included for the study if they did not have any hand injuries or hand-eye coordination problems. None of them had prior experience working with cobots. After obtaining their consent, they were given training on operating the robot program and performing the item sorting task in collaboration with the robot. The Institutional Review Board approved the research protocol at the University of Illinois at Chicago (IRB\# 2021-0936).

\subsection{Experimental procedure}
The experimental setup is shown in Figure 2. The goal of the sorting task was to separate the recyclable items, which are the red plastic cups, from the rest, which are non-recyclable (e.g., wood piece, scrap metal, plastic bags). The recyclable cups were laid out on a black foam board in a fixed circular arrangement. The non-recyclable items were randomly interspersed with the recyclables, but the layout was fixed for all participants. The participants were instructed to perform the task together with a cobot (UR3e, Universal Robots) equipped with a parallel clamping gripper (2FG7, OnRobot) and a vision system (Eyes, OnRobot). The participant's role was to make it easier for the robot’s vision system to detect and grip the cups. The study adopted a within-subjects factorial design where all participants completed all levels of the independent variables. 

\begin{figure}[htbp]
	\centering
	\vspace{-10pt}
	\includegraphics[width=0.9\linewidth]{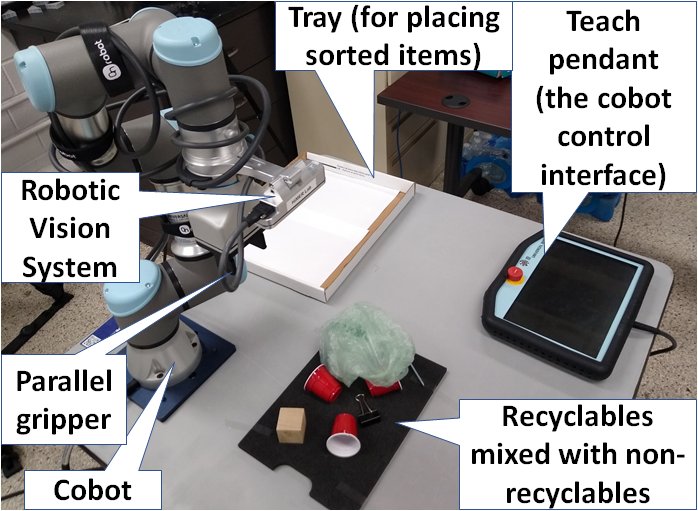}
	\vspace{-8pt}
	\caption{\small Experimental setup for the recyclable item sorting HRC task.}
	\label{fig:figure1}
    \vspace{-10pt}
\end{figure}

\vspace{0.2cm}

\textit{Independent variable \#1} (IV\#1, human involvement level): the extent of human involvement consisted of 3 levels in increasing order.
1) Level 1 (Occlusion removal): The subject must remove obstructions to the robot’s vision, like plastic bags that were on the cups.
2) Level 2 (Optimal spacing): The subject needs to remove occluding materials (= Level 1), and additionally separate the cups by a distance of at least 1 inch, so that there is sufficient gripper clearance for the robot to pick up the cups.
3) Level 3 (Optimal grip): The subject must remove occluding materials (= Level 1), space out the cups (= Level 2), then turn the cups upside down (bottom side up) so that the robot’s parallel gripper has a better grip on the cup during the pick and place movement. 
The examples of the outcomes after completing each human involvement level are shown in Figure 3. After the subject performed their role in the task, they were instructed to start the robot’s program on the teach pendant (the cobot control interface). The robot then sorted the cups into a separate tray.

\begin{figure}[htbp]
	\centering
	\vspace{-10pt}
	\includegraphics[width=1.0\linewidth]{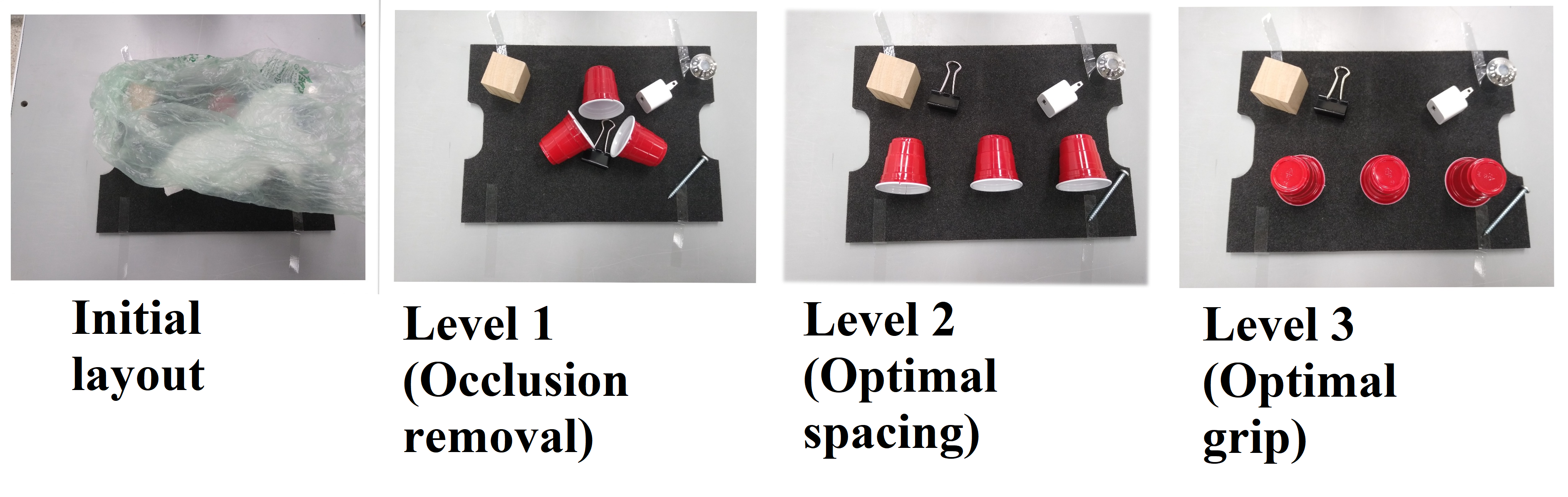}
	\vspace{-18pt}
	\caption{\small (Far left) The initial layout of the task board. (Left to right) the outcomes after completing each human involvement level, in increasing order.}
	\label{fig:figure3}
    \vspace{-10pt}
\end{figure}

\textit{Independent variable \#2} (IV\#2, percentage of recyclables: PR): the ratio of recyclables to the total number of items. Two different PRs were set up: PR = 30\% (3 items out of 10 items on the task board were recyclable) and PR = 70\% (7 out of 10 items were recyclable). Note that the 30\% PR is shown in Figure 3.
There are 6 experimental conditions (3 levels of IV\#1 $\times$ 2 levels of IV\#2). Since the sample size was small (n = 6), partial counterbalancing with Latin Square design was done for this study. Participants were randomly assigned to the different sequences. 
\vspace{-0.9mm}
\subsection{Measurements}
Performance metrics (task completion time, accuracy): Each session was video recorded. The time for human action, time for robot action, total task time, and robot accuracy were extracted from the videos to evaluate the task performance of the human-robot team. Accuracy is measured as the number of cups correctly sorted by the robot divided by the initial number of cups on the task board. Human time is calculated as the time between starting the human action (removing the occluding material) and starting the robot program on the teach pendant. Robot time is calculated as the time between the start of the robot movement to pick up the first cup and the placement of the last cup. Total time is the sum of the human time and robot time. 

Fluency and comfort: After each session, participants were required to fill out surveys assessing subjective fluency and comfort in working with the robot. Fluency refers to the level of coordination between a human and robot in a shared activity. A well-synchronized meshing of their actions indicates a high level of fluency \cite{Hoffman2019}. The survey on subjective fluency assessed the perceived coordination between the human and robot \cite{Hoffman2019}. It consisted of the following two items rated on a 5-point Likert scale (‘Strongly disagree = 1’ to ‘Strongly agree = 5’):
\begin{itemize}
\item “The robot and I worked fluently together to complete the task.”
\item “I felt that the actions of the robot were at the right timing.”
\end{itemize}

Similarly, the subject’s comfort level \cite{Olatunji2021} in working with the robot was assessed using the following two items, rated on the same 5-point Likert scale:

\begin{itemize}
\item “I felt comfortable working with the robot.”
\item “The experience with the robot made me stressed.” (reverse scored)
\end{itemize}

\subsection{Data Analysis}
 The average values of the outcome variables were calculated to compare the differences between the manipulated conditions. Two-way repeated measures analysis of variance was performed on each of the dependent variables to test the effects of the independent variables. The significance level is reported at $\alpha$ = 0.05. Post-hoc analysis with Bonferroni correction was performed to compare the three levels of the independent variable human involvement level.

\section{Results}

\subsection{Performance Metrics}

\begin{itemize}
\item \textbf{\textit{Accuracy:}} Accuracy for this task is the percentage of plastic cups correctly sorted by the robot. The accuracy was impacted by the human involvement level, which increased with increasing human involvement level. The accuracy was higher for Level 3 (optimal grip, mean accuracy = 100\%) compared to 2 (optimal spacing, mean accuracy = 69\%) and Level 1 (occlusion removal, mean accuracy = 33.3\%). The above results are shown in Figure 4.

\item \textbf{\textit{Human time:}} The human time for the task was impacted by the interaction between the two factors, human involvement level and percentage of recyclables, wherein the human time increased for 70\% recyclables (mean = 21s) as compared to 30\% recyclables (mean = 16s).

\item \textbf{\textit{Robot time:}} The percentage of recyclables affected the robot time, where the robot time was higher (mean = 65s) for 70\% recyclables compared to 30\% recyclables (mean = 28s). The robot time did not vary with human involvement level since it was programmed to perform the sorting task with a constant cycle time.

\vspace{-5pt}
\end{itemize}

\begin{figure}[htbp]
	\centering
	\vspace{-10pt}
	\includegraphics[width=1.0\linewidth]{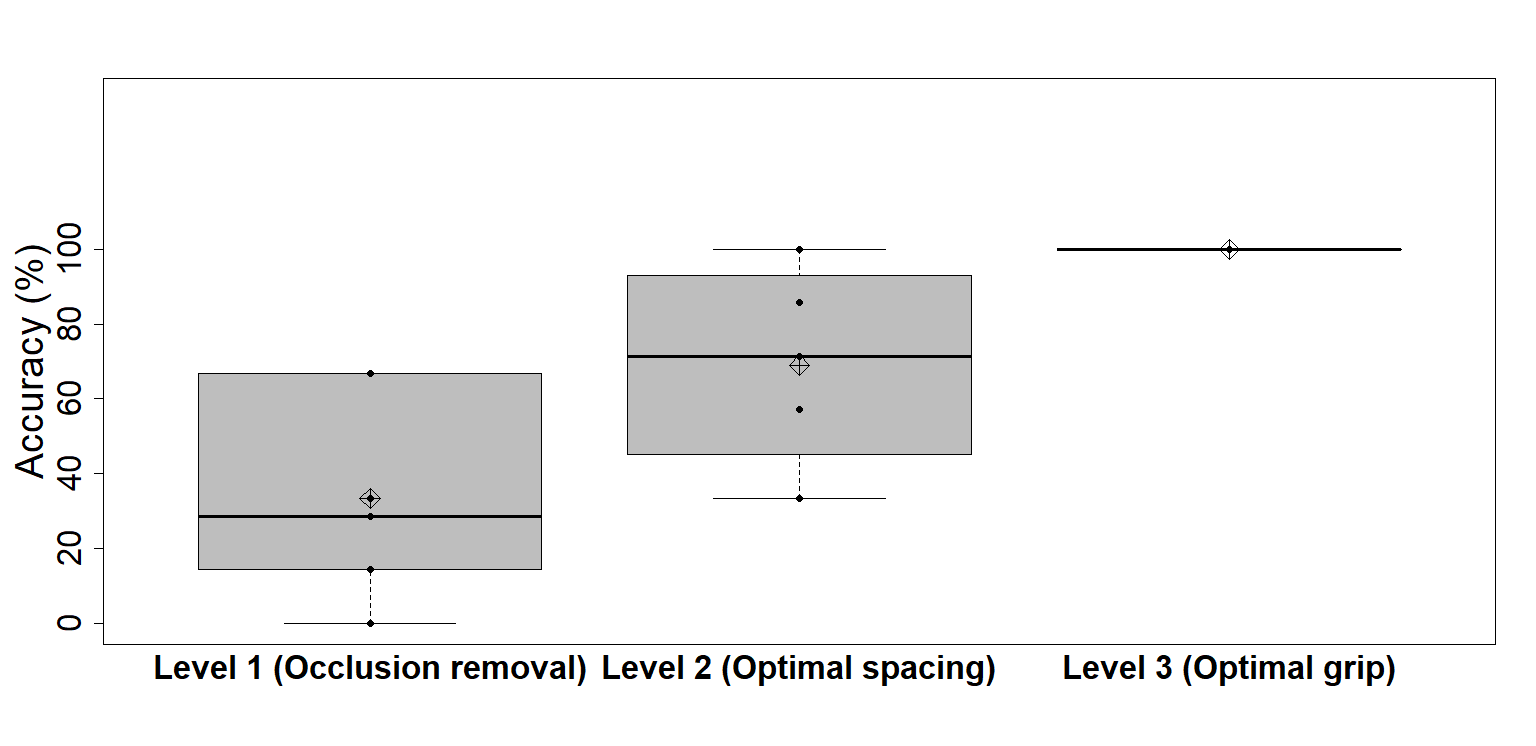}
	\vspace{-28pt}
	\caption{\small Variation of robot accuracy with different levels of human involvement.}
	\label{fig:figure4}
    \vspace{-10pt}
\end{figure}

\subsection{Subjective measures}
\begin{itemize}
\item \textbf{\textit{Fluency:}} Human involvement level affected subjective fluency. Fluency was higher for Level 3 (Optimal grip, mean = 9.6) compared to Level 2 (Optimal spacing, mean = 8.5) and Level 1 (Occlusion removal, mean = 8.1). The above results are shown in Figure 5.

\begin{figure}[htbp]
	\centering
	\vspace{-10pt}
	\includegraphics[width=1.0\linewidth]{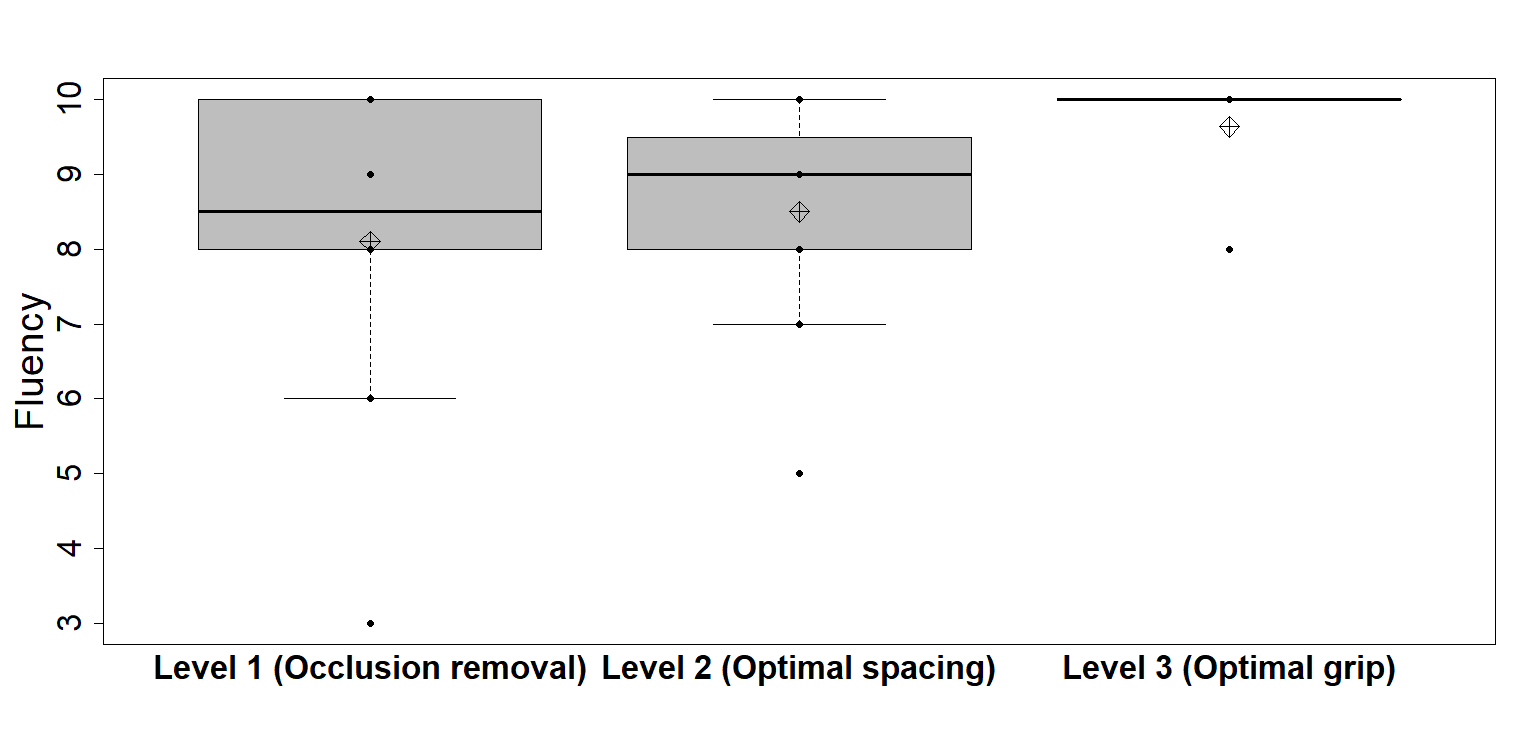}
	\vspace{-28pt}
	\caption{\small Variation of human-robot fluency with different levels of human involvement.}
	\label{fig:figure5}
    \vspace{-10pt}
\end{figure}

\item \textbf{\textit{Comfort:}} Neither the human involvement level nor percentage of recyclables had a remarkable effect on perceived comfort.
\end{itemize}

\section{Discussion}
The present study investigated the impact of human involvement level and percentage of recyclables on performance metrics such as task accuracy, task time, and subjective measures such as fluency and comfort in human-robot collaboration for a recycling sorting task. The main takeaways of this study are discussed as follows:

It was found that increasing the level of human involvement increased robot accuracy and the success of the human-robot team. The human’s actions made it easier for  the robot’s vision system to detect, pick and sort the plastic cups. The human’s actions of removing occluding materials, spacing out the objects and orienting the objects to the optimal gripping position increased the robot’s accuracy by providing the necessary gripper clearance and making it easier for the robot to pick up the items. Items on the recycling conveyor belts are randomly located and clumped together due to prior mechanical processes. Hence, it is challenging for fully automated sorting robots to grab them precisely \cite{Sarc2019}. In such scenarios, human assistance in spacing out the objects and orienting them to the optimal gripping position can improve the robot’s performance.

The perceived fluency increased with increase in human involvement. A possible explanation is that the subjects based their perception of fluency (or coordination between themselves and the robot) on the success of the human-robot team as a whole. It is likely that with higher human involvement when the robot was more accurate in sorting the recyclables, the subjects felt like they did a good job working with the robot and helping it achieve the team goal.

The findings of this experimental study have implications for realistic scenarios such as the sorting of complex materials that differ in material properties but appear visually similar, making it a challenge for a machine vision-based robot to identify and sort the materials all by itself accurately. Such tasks can be made more efficient with human input or assistance. For example, plant-based biodegradable plastics are increasingly becoming popular as a sustainable alternative to conventional petroleum-based plastics. They need to be sorted into a separate stream and sent to an industrial composting facility as they cannot be recycled like petroleum-based plastics. These bioplastics look identical to their petroleum-based counterparts in terms of color, size, thickness, and texture, making it challenging for robotic vision systems to identify and sort the materials accurately. On the other hand, with sufficient practice, human eyes and hands can be trained to identify the confounding materials and sort them more efficiently compared to a robot \cite{Guyot-Phung2019}.

\textit{Limitations:} 
The study is limited by an insufficient sample size. The results of this study are limited to the current experimental setup of the robot, that is, the type of gripper used (clamping type) and the composition of the waste stream (one type of recyclable item mixed with non-recyclables). If we had used a different gripper, such as a vacuum gripper, the robot would have been able to grip the object regardless of orientation. Hence, the robot accuracy would not have been very different for human involvement Levels 2 and 3. The human operator’s actions and human time could be different for a recycling scenario involving a different waste composition, such as a mix of different types of plastics such as PET, PVC, PP, etc. Furthermore, the setup differs from the recycling scenario in the real world. The objects are stationary, while the objects are on a moving conveyor belt in an actual recycling plant. 

\textit{Future work:} We plan to modify the task design with different types of plastics mixed in various proportions for our future work. In such a case, the human operator would have to use vision, touch, and learned experience to classify the plastics into different types. We also wish to modify the task design to include a component with the human and robot working simultaneously and task switching. Furthermore, we want to understand the physiological signatures of mental workload during the HRC task by analyzing features from biosignals, such as ECG.

\section*{Acknowledgment}

The authors thank Alayna Nguyen, Brian Martin, Matteo Capra, and Aju Thomas for their help with the experimental setup and data collection, and Ann Hsieh for her assistance with Figure 1.



\end{document}